\documentclass[sigconf]{acmart}

\setcopyright{acmlicensed}
\copyrightyear{2025}
\acmYear{2025}
\acmDOI{XXXXXXX.XXXXXXX}
\acmConference[ICVGIP '25]{ICVGIP 2025}{June 03--05,
  2025}{Mandi, India}

\acmISBN{978-1-4503-XXXX-X/2018/06}

\usepackage{natbib}
\usepackage{amsmath,amssymb,amsfonts}
\usepackage{algpseudocode}
\usepackage{graphicx}
\usepackage{xcolor}
\usepackage{booktabs} 
\usepackage{multirow} 
\usepackage{adjustbox} 
\usepackage{algorithm} 
\usepackage{subcaption}

\begin{document}

\title{SCoDA: Self-supervised Continual Domain Adaptation}

\author{Chirayu Agrawal}
\email{ca501@snu.edu.in}

\author{Snehasis Mukherjee}
\email{snehasis.mukherjee@snu.edu.in}
\affiliation{%
  \institution{Shiv Nadar Institution of Eminence}
  \city{Delhi NCR}
  \country{India}
}

\renewcommand{\shortauthors}{Agrawal et al.}

\begin{abstract}
Source-Free Domain Adaptation (SFDA) addresses the challenge of adapting a model to a target domain without access to the data of the source domain. Prevailing methods typically start with a source model pre-trained with full supervision and distill the knowledge by aligning instance-level features. However, these approaches, relying on cosine similarity over L2-normalized feature vectors, inadvertently discard crucial geometric information about the latent manifold of the source model. We introduce Self-supervised Continual Domain Adaptation (SCoDA) to address these limitations. We make two key departures from standard practice: first, we avoid the reliance on supervised pre-training by initializing the proposed framework with a teacher model pre-trained entirely via self-supervision (SSL). Second, we adapt the principle of geometric manifold alignment to the SFDA setting. The student is trained with a composite objective combining instance-level feature matching with a Space Similarity Loss. To combat catastrophic forgetting, the teacher's parameters are updated via an Exponential Moving Average (EMA) of the student's parameters. Extensive experiments on benchmark datasets demonstrate that SCoDA significantly outperforms state-of-the-art SFDA methods. The codes will be available after acceptance.
\end{abstract}



\received{20 February 2007}
\received[revised]{12 March 2009}
\received[accepted]{5 June 2009}

\maketitle

\section{Introduction}
\label{sec:introduction}
The success of Deep Learning in the computer vision area relies on vast, meticulously labeled datasets. However, a model's performance often drops when it is applied on a new environment (task) where the data distribution is different from its source environment - a problem known as domain shift. While Unsupervised Domain Adaptation (UDA) is the primary way to address the problem of domain shift, data privacy and intellectual property rules have created a tougher challenge: how to adapt a model to the target domain, without the knowledge learned from the source domain? This problem is called Source-Free Domain Adaptation (SFDA).
\begin{figure}[t]
    \centering
    \includegraphics[width=\columnwidth]{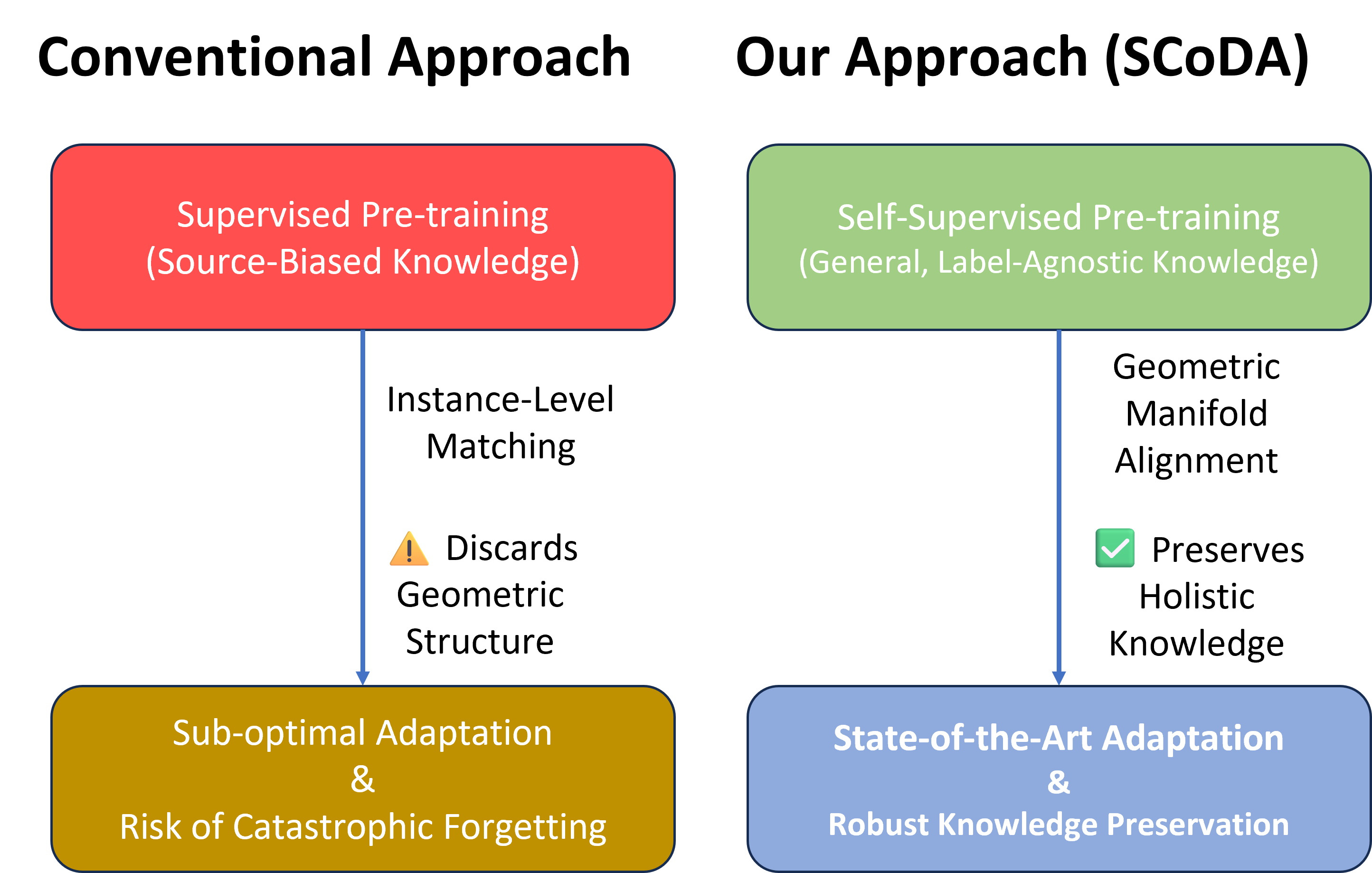}
    \caption{The core contribution of SCODA. We replace biased, supervised pre-training with a general, self-supervised foundation. We don't use brittle, instance-level feature matching because it discards crucial information. Instead, we use robust geometric manifold alignment to preserve the model's holistic knowledge. This leads to state-of-the-art adaptation while preventing catastrophic forgetting.}
    \label{fig:motivation}
\end{figure}

With SFDA, we adapt a pre-trained model using only unlabeled data from the target, which introduces a major risk of catastrophic forgetting, where the model loses its source knowledge. Most SFDA strategies use a ``teacher" model to distill knowledge to a ``student" \cite{singh2021unsupervised}. However, these methods have two key limitations. First, knowledge distillation approaches embed biases from the source classes, which hinders adaptation (thus compromising plasticity). Second, these approaches discard crucial geometric information about the teacher's learned knowledge space.

In the context of Unsupervised Knowledge Distillation (UKD), the typical reliance on aligning L2-normalized feature vectors is a critical flaw in the existing SFDA models, as it discards the holistic geometric structure of the teacher's learned manifold. The knowledge being transferred to the student is about \textit{what} the teacher represents for each sample, but not \textit{how} the teacher organizes its entire latent space.

Moreover, a majority of SFDA frameworks commence with a source model pre-trained with full supervision on a labeled source dataset \cite{yang2021gsfda}. This approach often embeds biases from the source classes into the model. That bias hinders adaptation when we move to target domains with different visual styles or class distributions. We posit that a model pre-trained via self-supervised learning (SSL) approach, which learns more general, label-agnostic visual representations, provides a more robust and adaptable foundation for SFDA.

To fix these problems, we propose Self-supervised Continual Domain Adaptation (SCODA). Our work makes a key contribution by adapting the principle of geometric manifold alignment from the Unsupervised Knowledge Distillation (UKD) literature to the SFDA setting. We integrate the \textbf{Space Similarity Loss} proposed by Singh et al.~\cite{singh2021unsupervised}, into a stable teacher-student framework inspired by recent work in continual adaptation such as CoSDA~\cite{feng2023cosda}. This dual-speed architecture provides a consistent learning target and effectively mitigates catastrophic forgetting. Figure \ref{fig:motivation} summarizes the proposed idea of SFDA in contrast with the traditional SFDA approaches. The main contributions of this paper are summarized as follows:
\begin{itemize}
    \item We update the standard SFDA paradigm by replacing conventional supervised source pre-training with a fully self-supervised one, demonstrating that label-agnostic representations provide a more effective starting point for domain adaptation.
    \item We are the first to adapt and validate the principle of geometric manifold alignment, specifically the Space Similarity loss from~\cite{singh2021unsupervised}, for the task of Source-Free Domain Adaptation.
    \item We integrate this geometric distillation into a dual-speed teacher-student architecture with a continual learning set up~\cite{feng2023cosda} to enable robust adaptation while mitigating catastrophic forgetting.
    \item Through extensive experiments, we show that SCoDA achieves state-of-the-art performance on the benchmark datasets, validating the efficacy of our approach.
\end{itemize}

Next, we conduct a survey on the related literature.

\section{Related Work}
\label{sec:related_work}
We start with a brief survey on the efforts made for unsupervised SFDA, followed by the shortcomings in such methods. Then we discuss a brief literature on Manifold alignment concept and the adaptation with continual learning setup, citing the motivation of our work.
\subsection{Unsupervised SFDA}
Unsupervised SFDA is a domain adaptation method where the goal is to adapt a pre-trained model to an unlabeled target domain without access to the knowledge obtained from the source domain data \cite{kim2021a2net}. SHOT \cite{liang2020we} freezes the classifier module (hypothesis) of the source model and learns the target-specific feature extraction module by exploiting both information maximization and self-supervised pseudo-labeling to implicitly align representations from the target domains to the source hypothesis. An improvement over SHOT was proposed in SHOT++~\cite{liang2022shotpp}, where a new label transfer strategy was introduced separating the target data into two splits based on the confidence of predictions (labelling information). Then they employ semi supervised learning approach to improve the accuracy of less-confident predictions in the target domain. A2Net~\cite{xie2021a2net} proposes an adaptive adversarial network to only access the well trained source model instead of source data during adaptation. Li et al.~\cite{li2023elr} focus on handling noisy pseudo-labels for SFDA. TPDS~\cite{tang2024tpds} introduces a flow of proxy distributions that facilitates the bridging of typically large distribution shift from the source domain to a target domain.

Fewer efforts are made for unsupervised SFDA. G-SFDA~\cite{yang2021gsfda} enables adaptation to an unlabeled target domain while preserving source performance, using local structure clustering for target alignment and sparse domain attention to retain source knowledge without accessing source data. Co-Learn and Co-learn++~\cite{zhang2022vlp} aim to adapt a source model trained on a fully labeled source domain to a related but unlabeled target domain by proposing a framework to incorporate pre-trained networks into the target adaptation process. AaD~\cite{yang2022attracting} encourages target features to move closer to their corresponding class prototypes while pushing them away from other classes, effectively modeling class structure in the feature space.

Efforts have been made to apply contrastive learning approaches to train the model on the target domain, avoiding the source domain information. DaC~\cite{zhang2022dac} introduces adaptive contrastive learning by splitting the target domain into confident and uncertain regions, enabling the model to learn discriminative features from both parts. AdaCon~\cite{chen2022adacon} employs a contrastive test-time adaptation objective that enforces feature consistency across multiple augmented views of the same target sample. SFDA-DE~\cite{ding2022sfda-de} models the target data distribution using a Gaussian Mixture Model in the feature space, aiming to produce more accurate pseudo-labels for self-training. CoWA~\cite{lee2022cowa} proposes a collaborative weighting strategy that combines model predictions with the structure of feature-space clusters to compute more reliable pseudo-label confidence scores.

\subsection{Shortcomings of the Unsupervised SFDA methods} 
Most existing methods rely on the pseudo labeling of data and confidence thresholds. This may lead to noisy pseudo-label propagation under domain shift, and may consequently lead to catastrophic forgetting due to loss of source domain knowledge because of the lack of regularization or structure preservation. These methods also lead to instability due to unreliable targets and weak feature anchoring. The methods support only instance-level alignment and ignore the global geometry of feature space. Most importantly, these methods require a teacher model trained using supervised learning for domain adaptation.

The proposed SCoDA improves on this by using a self supervised model as a teacher model. This model learns more general and label-agnostic features which provide a robust starting point. SCoDA preserves the teacher model's latent feature geometry using a space similarity loss which ensures that the global structure of the learned representation is maintained during adaptation. The EMA-based teacher-student framework helps to enable smooth and stable adaptation preventing catastrophic forgetting.

\subsection{Manifold Alignment in Knowledge Distillation}
The concept of preserving the teacher's latent geometry is a powerful but relatively new idea in knowledge transfer. Miles et al. \cite{vkdcvpr2024} propose a constrained feature distillation method containing two components: an orthogonal projection and a task specific normalisation. Hi et al.~\cite{topoguided2025} introduce a knowledge distillation framework that leverages topology-aware representations and gradient-guided knowledge distillation to effectively transfer knowledge from a high-capacity teacher to a lightweight student model. Singh et al.~\cite{singh2021unsupervised} identified that aligning $L_2$-normalized features - a common practice in the literature - destroys the teacher's learned manifold structure. They proposed a ``space similarity" loss to fix this for Unsupervised Knowledge Distillation (UKD).

Most of the conventional knowledge distillation methods are based on aligning the logits or L2-normalized features between teacher and student models. These methods, although simple, have the side effect of losing the teacher's manifold structure of learning, as relational context is collapsed to local point-wise similarity through feature normalization. Local similarity assumption hinders the student from capturing the teacher's global geometric and topological structure, which is most important for learning tasks that are based on more structured, richer representations. Therefore, these methods tend to learn weaker student models that generalize poorly to challenging or out-of-distribution data where relational context preservation is critical. SCoDA makes a crucial contribution by proposing a framework to adapt, validate, and apply the principle of geometric manifold alignment by using both cosine loss and space similarity loss to the SFDA problem, where the model must bridge a significant domain gap \textit{without} any source data.

\subsection{Teacher–Student Frameworks for Continual Adaptation}
Tarvainen et al.~\cite{tarvainen2017mean} averages the teacher model weights instead of label predictions which improved semi supervised results. Teacher Adaptation \cite{szatkowski2024adapt} concurrently updates the teacher and the main models during incremental training. Using a teacher-student setup, where the teacher's weights are an Exponential Moving Average (EMA) of the student's, is a proven technique for stabilizing training. This architecture has been instrumental in continual adaptation. CoSDA \cite{feng2023cosda} used this dual-speed framework to prevent catastrophic forgetting during adaptation to a sequence of target domains.

While these frameworks are effective at stabilizing training, their mechanism for knowledge transfer is fundamentally limited. CoSDA and its successors still primarily distill knowledge by matching instance-level features (usually via cosine similarity on normalized vectors). As we have argued, this discards vital geometric information about the teacher's manifold. The proposed SCoDA improves upon this by adopting the stable EMA teacher-student architecture but revolutionizing \textit{what} knowledge is transferred. We replace the simple instance-level consistency loss with a more powerful composite objective featuring the Space Similarity Loss. This ensures the student learns not just what the teacher thinks about one sample, but \textit{how} the teacher organizes its entire latent space, leading to a much more robust and effective adaptation. Next, we illustrate the proposed SFDA method.
\begin{figure*}[ht!]
    \centering
    \includegraphics[width=\textwidth,height=4in]{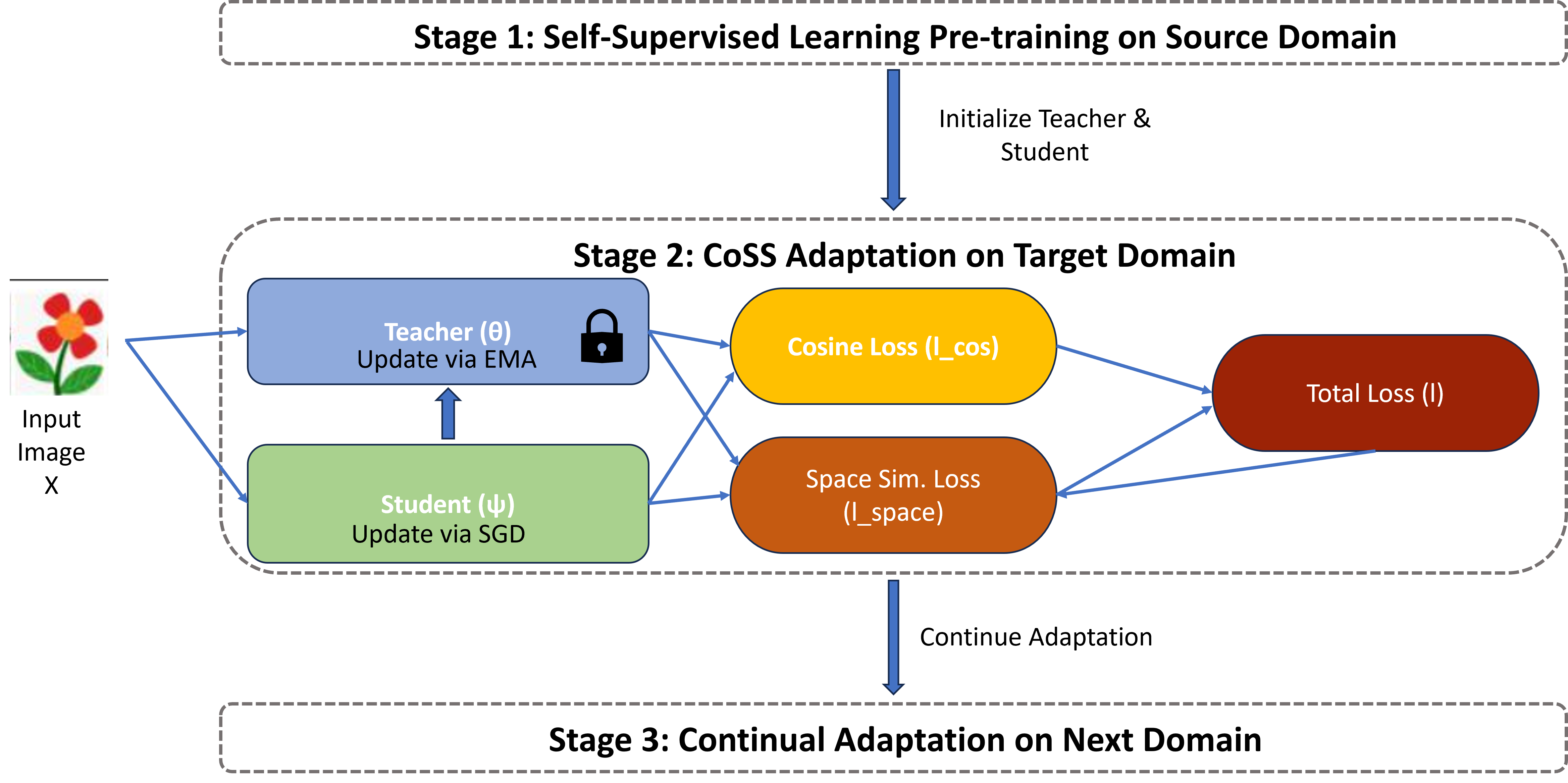}
    \caption{A detailed schematic of the proposed SCoDA framework. \textbf{Stage 1 (Initialization):} The process begins by leveraging a powerful model pre-trained on the source domain using an SSL objective, departing from standard supervised pre-training. Its weights are used to initialize an identical teacher ($\theta$) and student ($\psi$) network. \textbf{Stage 2 (Target Domain Adaptation):} This is the core adaptation loop, inspired by continual learning frameworks~\cite{feng2023cosda}. For each unlabeled input image, the student model actively learns via SGD. The supervisory signal comes from the teacher model, driven by a composite loss function adapting the CoSS objective from~\cite{singh2021unsupervised}: the Cosine Similarity Loss ($\mathcal{L}_{\text{cos}}$) for instance-level consistency and the Space Similarity Loss ($\mathcal{L}_{\text{space}}$) for structural consistency. The teacher model evolves slowly as an EMA of the student's parameters, providing a stable learning target. \textbf{Stage 3 (Continual Adaptation):} The fully adapted module can be seamlessly used to initialize adaptation on a new, subsequent target domain.}
    \label{fig:scoda_architecture_detailed}
\end{figure*}

\section{Proposed Method}
\label{sec:method}
We start with an illustration of the overall proposed framework, followed by a detailed discussion on each of our contributions.
\subsection{Framework Overview}
We propose SCoDA, a framework designed to perform a complete knowledge transfer in the source-free setting. The overall pipeline is illustrated in Figure~\ref{fig:scoda_architecture_detailed}.

\subsubsection{Stage 1: Initialization} A key departure from conventional SFDA lies in our initialization strategy. Instead of using a source model pre-trained with supervision, we begin with a robust model $h$ pre-trained on the source domain using a self-supervised learning (SSL) objective, such as BYOL~\cite{grill2020bootstrap}. This provides a rich, label-agnostic manifold as a starting point. Both the teacher model $f_\theta$ and the student model $f_\psi$ are initialized with the weights of this source model.

\subsubsection{Stage 2: Iterative Adaptation} The framework then iteratively adapts to the unlabeled target domain $D_T$. In each training step, a mini-batch of data $X$ is sampled from $D_T$. The student model $f_\psi$ is updated via gradient descent on a composite loss function guided by the teacher model $f_\theta$. Following the student update, the teacher's parameters $\theta$ are updated as an EMA of the student's new parameters $\psi$.
\begin{figure*}[t!]
    \centering
    \includegraphics[width=\textwidth]{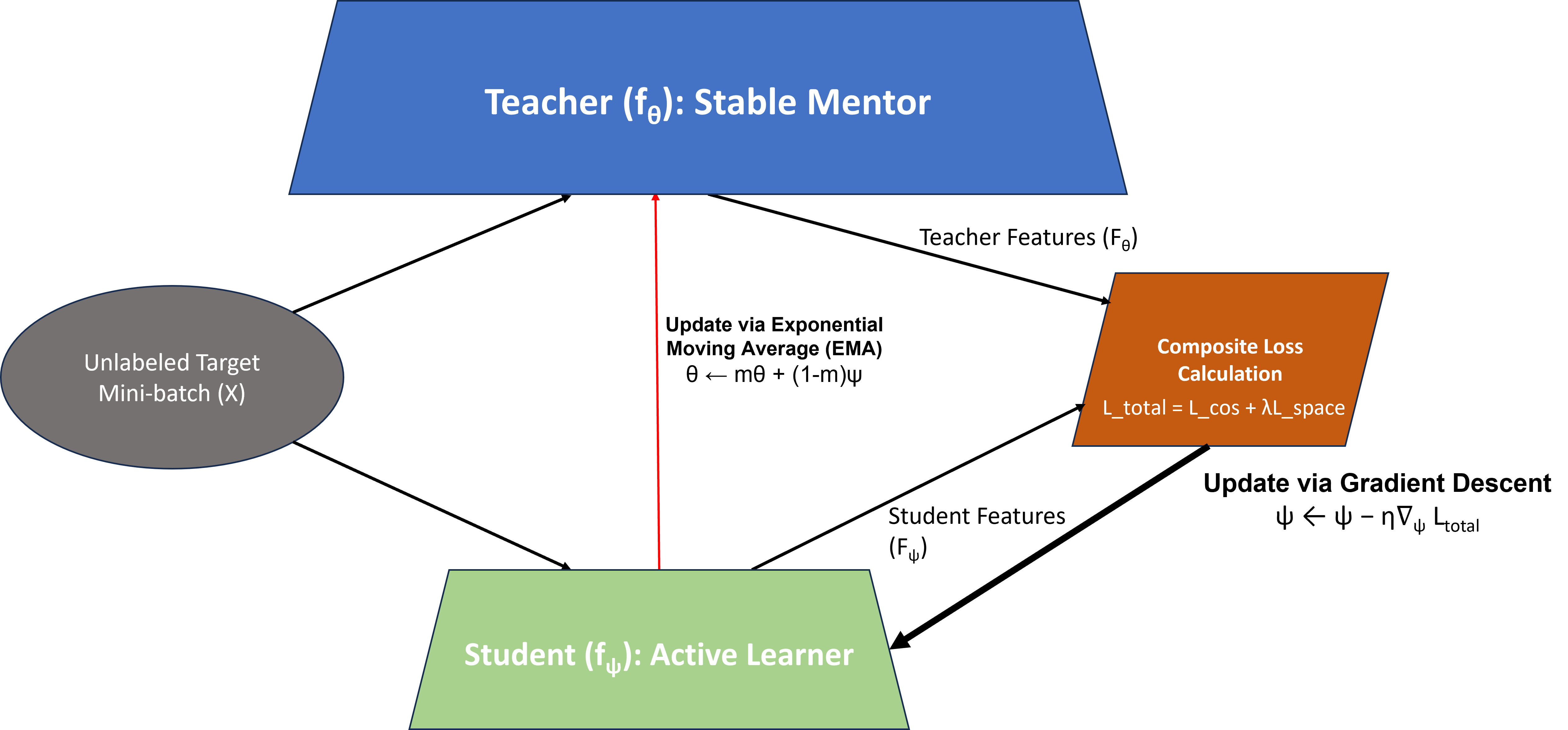}
    \caption{A detailed view of the Teacher-Student adaptation dynamics in SCODA.An input mini-batch (X) is processed by both the student ($f_\psi$) and the teacher ($f_\theta$).The student is actively trained via gradient descent on a composite loss ($\mathcal{L}_\text{total}$).The teacher provides a stable supervisory signal and is updated as a slow-moving average (EMA) of the student's weights, which mitigates catastrophic forgetting.}
    \label{fig:teacher_student_loop}
\end{figure*}

\subsection{Adapting Geometric Manifold Alignment for SFDA}
A central contribution of our work is the adaptation of the geometric distillation method from~\cite{singh2021unsupervised} to the SFDA problem. Relying only on cosine similarity over L2-normalized features is flawed because normalization is not a homeomorphism and discards crucial geometric information. To address this problem, we adopt the composite loss.

Let a mini-batch of target data be $X \in \mathbb{R}^{B \times C \times H \times W}$, where $B$ is the batch size. The teacher and student models process this batch to produce feature matrices $F_\theta, F_\psi \in \mathbb{R}^{B \times d}$, where $d$ is the feature dimension. Our composite loss consists of two components:

\subsubsection{Feature Similarity Loss ($\mathcal{L}_{\text{cos}}$)}
This standard term aligns individual feature vectors by maximizing their cosine similarity.
\begin{equation}
\mathcal{L}_{\text{cos}} = 1 - \frac{1}{B} \sum_{i=1}^{B} \frac{F_{\theta,i} \cdot F_{\psi,i}}{\|F_{\theta,i}\|_2 \|F_{\psi,i}\|_2}.
\end{equation}
In this equation, $F_{\theta,i}$ and $F_{\psi,i}$ are the feature vectors from the teacher and student models, respectively, for the $i$-th sample in the batch. The term $\|\cdot\|_2$ denotes the L2-norm.

\subsubsection{Space Similarity Loss ($\mathcal{L}_{\text{space}}$)}
To capture feature-space geometry, we use the Space Similarity loss from \cite{singh2021unsupervised}. We transpose the feature matrices to $F_\theta^T, F_\psi^T \in \mathbb{R}^{d \times B}$ and define the loss as:
\begin{equation}
\mathcal{L}_{\text{space}} = 1 - \frac{1}{d} \sum_{j=1}^{d} \frac{F_{\theta,j}^T \cdot F_{\psi,j}^T}{\|F_{\theta,j}^T\|_2 \|F_{\psi,j}^T\|_2}.
\end{equation}
Here, $F_{\theta,j}^T$ and $F_{\psi,j}^T$ are the $j$-th rows of the transposed teacher and student feature matrices, respectively. Each row represents the activation values of a specific feature dimension across the entire mini-batch. Minimizing $\mathcal{L}_{\text{space}}$ ensures structural consistency across dimensions, preserving the latent space’s internal grammar.

\subsection{Stabilizing Adaptation via Teacher-Student Dynamics}
To address catastrophic forgetting, we employ a dual-speed teacher-student framework inspired by continual adaptation method CoSDA~\cite{feng2023cosda}. The teacher model $f_\theta$ is a slowly evolving EMA of the student, providing a stable learning target.

The total loss function, $\mathcal{L}_{\text{total}}$, is a weighted sum of the two component losses:
\begin{equation}
\mathcal{L}_{\text{total}} = \mathcal{L}_{\text{cos}} + \lambda \mathcal{L}_{\text{space}},
\end{equation}
where $\lambda$ is a hyperparameter that balances the contribution of the two loss components. The models' parameters are updated iteratively. Let $\psi$ and $\theta$ denote the parameters of the student and teacher models, respectively. For each training iteration $t$, the following updates are performed:

\subsubsection{Student Update}
The student parameters $\psi_t$ at iteration $t$ are updated by gradient descent:
\begin{equation}
\psi_{t+1} \leftarrow \psi_t - \eta \nabla_\psi \mathcal{L}_{\text{total}}(F_{\theta_t}(X_t), F_{\psi_t}(X_t)),
\end{equation}
where $\psi_{t+1}$ are the parameters after the update, $\eta$ is the learning rate, and the gradient $\nabla_\psi$ is computed with respect to the student parameters for the current mini-batch $X_t$.

\subsubsection{Teacher Update}
The teacher's parameters $\theta_t$ are updated as an exponential moving average (EMA) of the student's newly updated parameters:
\begin{equation}
\theta_{t+1} \leftarrow m \theta_t + (1 - m) \psi_{t+1},
\end{equation}
where $\theta_{t+1}$ are the updated teacher parameters and $m$ is the EMA momentum coefficient. The teacher’s BatchNorm statistics are also updated via EMA. The full procedure is detailed in Algorithm~\ref{alg:scoda}. Next, we discuss the experimental setup in detail.
\begin{algorithm}[ht!]
\caption{SCoDA: Self-supervised Continual Domain Adaptation}
\label{alg:scoda}
\begin{algorithmic}[1]
\State \textbf{Input:} Unlabeled target dataset $D_T$, SSL-pretrained model $h$, EMA momentum $m$, learning rate $\eta$, loss weight $\lambda$.
\State \textbf{Output:} Adapted student model $f_\psi$.
\State Initialize teacher $\theta \gets h$, student $\psi \gets h$.
\For{each mini-batch $X \in D_T$}
    \State Extract features: $F_\theta = f_\theta(X)$, $F_\psi = f_\psi(X)$.
    \State Compute $\mathcal{L}_{\text{cos}}$ using Eq. (1).
    \State Compute $\mathcal{L}_{\text{space}}$ using Eq. (2), adapted from~\cite{singh2021unsupervised}.
    \State Compute total loss: $\mathcal{L}_{\text{total}} = \mathcal{L}_{\text{cos}} + \lambda \mathcal{L}_{\text{space}}$.
    \State Update student: $\psi \gets \psi - \eta \nabla_\psi \mathcal{L}_{\text{total}}$.
    \State Update teacher: $\theta \gets m\theta + (1 - m)\psi$, inspired by~\cite{feng2023cosda}.
    \State Update teacher's BatchNorm statistics via EMA.
\EndFor
\end{algorithmic}
\end{algorithm}

\section{Datasets and Experiments}
\label{sec:experiments}
We first describe the datasets used for experiments, followed by experimental setup.

\subsection{Datasets}
We evaluate SCoDA on two standard benchmarks: \textbf{Office-Home}~\cite{venkateswara2017deep} and \textbf{DomainNet}~\cite{peng2019moment} datasets.

The \textbf{Office-Home} dataset~\cite{venkateswara2017deep} is a challenging benchmark specifically designed for domain adaptation tasks. It comprises approximately 15,500 images categorized into 65 distinct classes, spanning four visually distinct domains: Art, Clipart, Product, and Real-World. The images within this dataset exhibit significant domain shift due to variations in style, object appearance, and background clutter across these domains. Image resolutions vary but are typically around $256 \times 256$ pixels. The complexity of Office-Home lies in its fine-grained categories and the substantial discrepancy between domains, making it a robust testbed for evaluating cross-domain generalization capabilities.

The \textbf{DomainNet}~\cite{peng2019moment} dataset is a large-scale dataset for domain generalization and domain adaptation, offering a more extensive and complex evaluation setting. It contains approximately 345,000 images distributed across 345 classes and 34 diverse domains. Each domain represents a unique style or rendering of the same set of object classes, ranging from real-world photographs to sketches, paintings, and cartoon styles. Image resolutions also vary widely across the different domains and individual images. The sheer number of images, classes, and especially the large number of distinct domains contribute to the high complexity of DomainNet, posing a significant challenge for models to learn domain-invariant features and generalize effectively across unseen domains.

\subsection{Experimental Setup}
We follow the standard SFDA protocol for evaluation of the proposed approach. We compare SCoDA against the state-of-the-art baselines such as \textbf{Source-Only}, \textbf{SHOT}~\cite{liang2020we}, \textbf{NRC}~\cite{yang2021exploiting}, and \textbf{AaD}~\cite{yang2022attracting}. We use a \textbf{ResNet-50} backbone pretrained on ImageNet. The student is trained with SGD (momentum 0.9, weight decay $10^{-3}$, batch size 64). For SCoDA, we set $\lambda=1.0$ and $m=0.999$. The experiments were conducted on an NVIDIA RTX 4000 GPU. Next, we discuss the results of our experiments.

\section{Results and Discussions}
We first present the results, ablation studies followed by the detailed analysis of the results.

\subsection{Comparison Against the Competing Methods}
Table~\ref{tab:officehome_results} presents the classification accuracy of SCoDA on the Office-Home dataset, compared to the state-of-the-art. SCoDA achieves substantial improvements across all three task configurations (from source task Ar to target tasks Cl, Pr and Rw) evaluated in this study. Table~\ref{tab:domainnet_results} shows the classification accuracy of the proposed approach on the DomainNet dataset (from I to C task), demonstrating SCoDA's effectiveness on large-scale domain adaptation.

\begin{table}[ht!]
\caption{Classification Accuracy (\%) on Office-Home for SFDA. The \textbf{SCoDA (pre-adaptation)} row serves as the direct baseline for our method. The \textbf{SCoDA (post-adaptation)} row shows the final performance.}
\label{tab:officehome_results}
\begin{adjustbox}{center}
\begin{tabular}{@{}lcccc@{}}
\toprule
\textbf{Method} & \textbf{Ar→Cl} & \textbf{Ar→Pr} & \textbf{Ar→Rw} & \textbf{Average} \\
\midrule
Source-Only & 34.9 & 50.0 & 58.0 & 47.63 \\
\midrule
ZSC & 76.8 & 91.1 & 88.3 & 85.40 \\
A2Net~\cite{xie2021a2net} & 58.4 & 79.0 & 82.4 & 73.27 \\
SHOT~\cite{liang2020we} & 56.7 & 77.9 & 80.6 & 71.73 \\
SHOT++~\cite{liang2022shotpp} & 57.9 & 79.7 & 82.5 & 73.37 \\
CPGA~\cite{zhou2021cpga} & 59.3 & 78.1 & 79.8 & 72.40 \\
GKD~\cite{tang2022gkd} & 56.5 & 78.2 & 81.8 & 72.17 \\
NRC~\cite{yang2021exploiting} & 57.7 & 80.3 & 82.0 & 73.33 \\
AaD~\cite{yang2022attracting} & 59.3 & 79.3 & 82.1 & 73.57 \\
DaC~\cite{zhang2022dac} & 59.1 & 79.5 & 81.2 & 73.27 \\
AdaCon~\cite{chen2022adacon} & 47.2 & 75.1 & 75.5 & 65.93 \\
SFDA-DE~\cite{ding2022sfda-de} & 59.7 & 79.5 & 82.4 & 73.87 \\
CoWA~\cite{lee2022cowa} & 56.9 & 78.4 & 81.0 & 72.10 \\
SCLM~\cite{tang2022sclm} & 58.2 & 80.3 & 81.5 & 73.33 \\
ELR~\cite{li2023elr} & 58.4 & 78.7 & 81.5 & 72.87 \\
TPDS~\cite{tang2024tpds} & 59.3 & 80.3 & 82.1 & 73.90 \\
SHOT+DPC~\cite{xia2024dpc} & 59.2 & 79.8 & 82.6 & 73.87 \\
AaD+DPC~\cite{xia2024dpc} & 59.5 & 80.6 & 82.9 & 74.33 \\
AaD w/DCPL~\cite{diamant2024dcpl} & 61.9 & 84.5 & 87.7 & 78.03 \\
SHOT w/DCPL~\cite{diamant2024dcpl} & 59.9 & 82.3 & 87.8 & 76.67 \\
SHOT++ w/DCPL~\cite{diamant2024dcpl} & 61.2 & 84.3 & 88.0 & 77.83 \\
Improved SFDA~\cite{mitsuzumi2024improved} & 60.7 & 78.9 & 82.0 & 73.87 \\
HRD++~\cite{xing2024hrd} & 63.6 & 83.6 & 85.4 & 77.53 \\
SHOT w/ Co-learn~\cite{zhang2022vlp} & 63.1 & 83.1 & 84.6 & 76.93 \\
SHOT++ w/ Co-learn~\cite{zhang2022vlp} & 63.6 & 83.6 & 84.8 & 77.33 \\
NRC w/ Co-learn~\cite{zhang2022vlp} & 72.2 & 87.6 & 88.4 & 82.73 \\
AaD w/ Co-learn~\cite{zhang2022vlp} & 66.4 & 85.3 & 87.0 & 79.57 \\
ZSC w/ Co-learn~\cite{zhang2022vlp} & 77.2 & 90.4 & 91.0 & 86.20 \\
SHOT w/ Co-learn++~\cite{zhang2022vlp} & 62.2 & 83.7 & 84.8 & 76.90 \\
SHOT++ w/ Co-learn++~\cite{zhang2022vlp} & 62.5 & 83.5 & 84.5 & 76.83 \\
NRC w/ Co-learn++~\cite{zhang2022vlp} & 76.4 & 88.8 & 88.6 & 84.60 \\
AaD w/ Co-learn++~\cite{zhang2022vlp} & 71.9 & 88.1 & 86.7 & 82.23 \\
ZSC w/ Co-learn++~\cite{zhang2022vlp} & 80.0 & 91.2 & 91.8 & 87.67 \\
DIFO~\cite{tang2024difo} & 70.6 & 90.6 & 88.8 & 83.33 \\
RLD w/ NBF~\cite{song2024rld} & 62.2 & 81.0 & 79.7 & 74.30 \\
LFTL~\cite{lyu2024lftl} & 76.6 & 92.2 & 89.7 & 86.17 \\
\midrule
CoSDA~\cite{feng2023cosda} & 54.75 & 75.44 & 80.65 & 70.28 \\
CoSDA (+) AaD~\cite{feng2023cosda} & 55.21 & 74.70 & 80.65 & 70.19 \\
\midrule
\textbf{SCoDA (pre-adaptation)} & 53.56 & 89.05 & 81.78 & 74.80\\
\textbf{SCoDA (post-adaptation)} & \textbf{75.60} & \textbf{93.62} & \textbf{90.54} & \textbf{86.59} \\
\bottomrule
\end{tabular}
\end{adjustbox}
\end{table}
\begin{table}[ht!]
\caption{Classification Accuracy (\%) on the DomainNet I→C Task.}
\label{tab:domainnet_results}
\centering
\begin{adjustbox}{center}
\begin{tabular}{@{}lc@{}}
\toprule
\textbf{Method} & \textbf{Accuracy (\%)} \\ \midrule
Source-Only (Vanilla) & 13.43 \\
SHOT~\cite{warrad2023revisiting} & 14.82 \\
NRC~\cite{warrad2023revisiting} & 18.01 \\
AaD~\cite{warrad2023revisiting} & 47.12 \\
CoSDA~\cite{feng2023cosda} & 40.97 \\
CoSDA (+) AaD~\cite{feng2023cosda} & 46.42 \\ \midrule
\textbf{SCoDA (Ours)} & \textbf{37.68} \\ \bottomrule
\end{tabular}
\end{adjustbox}
\end{table}

\subsection{Ablation Studies}
We have performed several ablation studies to observe the efficacy of the proposed method. We have experimented with different state-of-the-art Self Supervised Learning (SSL) models with SCoDA. Table~\ref{tab:ssl_ablation} presents the performance of SCoDA across different SSL pre-training initializations on the Office-Home dataset (from Ar to Cl task).
\begin{table}[ht!]
\caption{Impact of Different SSL Pre-training Models on Office-Home (Ar→Cl).}
\label{tab:ssl_ablation}
\centering
\begin{tabular}{@{}lccc@{}}
\toprule
\textbf{SSL Model} & \textbf{Pre-Adapt (\%)} & \textbf{Post-Adapt (\%)} & \textbf{Gain ($\Delta$)} \\ \midrule
\textbf{BYOL} & \textbf{53.56} & \textbf{75.60} & \textbf{+22.04} \\
SimCLR & 13.00 & 23.40 & +10.40 \\
Barlow Twins & 4.47 & 5.27 & +0.80 \\
DINO & 48.00 & 51.32 & +3.32 \\ \bottomrule
\end{tabular}
\end{table}

Further, we perform an analysis on the amount of forgetting of the proposed method. Table~\ref{tab:forgetting_analysis} quantifies the target domain improvement and source domain retention capabilities of SCoDA on DomainNet. This shows the balance between plasticity and stability of the proposed approach.
\begin{table}[ht!]
\caption{Analysis of Adaptation Gain and Forgetting Mitigation on DomainNet.}
\label{tab:forgetting_analysis}
\centering
\begin{adjustbox}{width=\columnwidth,center}
\begin{tabular}{@{}lccc@{}}
\toprule
\textbf{Task Description} & \textbf{Task} & \textbf{Pre-Adapt (\%)} & \textbf{Post-Adapt (\%)} \\ \midrule
Target Domain Adaptation & I → C & 21.15 & 37.68 \\
Source Domain Forgetting & I → I & 16.85 & 16.74 \\ \bottomrule
\end{tabular}
\end{adjustbox}
\end{table}

We also experimented on effectiveness of the different components of the proposed loss function. Table~\ref{tab:loss_ablation} presents the results of the ablation study examining the contribution of each loss component to SCoDA. Performance enhancements are visible for introducing each component of the loss function.
\begin{table}[ht!]
\caption{Ablation Study of Loss Components on Office-Home (Ar→Cl).}
\label{tab:loss_ablation}
\centering
\begin{adjustbox}{center}
\begin{tabular}{@{}lc@{}}
\toprule
\textbf{Method/Configuration} & \textbf{Accuracy (\%)} \\ \midrule
\textbf{SCoDA (full: $\mathcal{L}_{\text{cos}} + \lambda \mathcal{L}_{\text{space}}$)} & \textbf{75.60} \\
SCoDA w/o $\mathcal{L}_{\text{space}}$ (Cosine only) & 73.20 \\
SCoDA w/o $\mathcal{L}_{\text{cos}}$ (Space only) & 72.40 \\ \bottomrule
\end{tabular}
\end{adjustbox}
\end{table}

\subsection{Result Analysis}

\subsubsection{Office-Home Benchmark Analysis}
SCoDA demonstrates competitive performance on the Office-Home benchmark, achieving an average accuracy of \textbf{86.59\%} across the three evaluated tasks. This represents a substantial improvement of \textbf{+11.79} percentage over the pre-adaptation baseline (74.80\%). The method shows particularly strong performance on tasks originating from the Art domain, with Ar→Pr achieving \textbf{93.62\%} and Ar→Rw reaching \textbf{90.54\%}. 

However, the results reveal that SCoDA falls short of the state-of-the-art ViLAaD++ method, which achieves \textbf{90.23\%} average accuracy. The performance gap is most pronounced on the Ar→Cl task, where SCoDA achieves 75.60\% compared to ViLAaD++'s 83.3\%. This suggests that while SCoDA's geometric alignment approach is effective, there is still room for improvement in handling the most challenging domain shifts, particularly those involving artistic to clipart transformations.

\subsubsection{DomainNet Benchmark Analysis}
On the DomainNet I→C task, SCoDA achieves \textbf{37.68\%} accuracy, which represents a significant improvement over basic methods such as Source-Only (13.43\%) and SHOT (14.82\%). However, the results reveal an interesting discrepancy: while SCoDA outperforms most baseline methods, it falls short of the AaD baseline (47.12\%) by approximately 9.44 percentage. This suggests that SCoDA's effectiveness may be more pronounced on certain types of domain shifts but may struggle with the specific characteristics of the Infograph→Clipart transformation in DomainNet.

The forgetting analysis provides crucial insights into SCoDA's stability. The method achieves a remarkable target domain improvement of \textbf{+16.53} percentage (from 21.15\% to 37.68\%) while maintaining exceptional source domain performance with only a negligible decrease of \textbf{-0.11\%} (from 16.85\% to 16.74\%). This near-perfect preservation of source domain knowledge validates SCoDA's ability to prevent catastrophic forgetting, which is a critical advantage in practical domain adaptation scenarios.

\subsubsection{Self-Supervised Learning Foundation Analysis}
The comparison of different SSL pre-training models reveals significant insights about the compatibility of SCoDA with various self-supervised approaches. BYOL emerges as the most effective foundation, providing not only the strongest initial accuracy (53.56\%) but also yielding the largest adaptation gain of \textbf{+22.04} percentage. This substantial improvement suggests that BYOL's learned manifold structure is particularly amenable to SCoDA's geometric alignment strategy.

In stark contrast, DINO shows the most limited improvement, with only a \textbf{+3.32} percentage gain despite starting from a relatively strong baseline (48.00\%). This poor performance aligns with the technical analysis revealing fundamental incompatibilities between SCoDA's adaptation approach and DINO's learning principles. The DINO analysis reveals severe catastrophic forgetting, with source domain performance dropping catastrophically from 48.00\% to 4.00\% after adaptation, while target domain performance remains extremely poor at 0.29\%.

SimCLR shows moderate performance with a +10.40 gain, though from a much lower baseline (13.00\%), while Barlow Twins demonstrates minimal improvement (+0.80), suggesting fundamental incompatibility with SCoDA's approach.

\subsubsection{Loss Function Component Analysis}
The ablation study of the loss components provides clear evidence for the complementary nature of SCoDA's composite loss design. The complete framework (75.60\%) significantly outperforms configurations using only cosine similarity (73.20\%) or only spatial structure loss (72.40\%). This analysis confirms that both loss components capture complementary aspects of the teacher's knowledge. The cosine similarity loss appears to provide stronger baseline performance, while the spatial structure loss contributes additional refinement. The fact that neither component alone achieves the performance of the whole method validates the theoretical motivation for the composite loss design.
\begin{table}[ht!]
\caption{DINO Model Performance on Office-Home (Art→Clipart).}
\label{tab:dino_performance}
\centering
\begin{tabular}{@{}p{0.2\columnwidth}cc@{}}
\toprule
\textbf{Model Configuration} & \textbf{Source Domain (Art)} & \textbf{Target Domain (Clipart)} \\ \midrule
Pre-trained DINO (on Art) & 48.00 & 15.19 \\
Adapted DINO (on Clipart) & 4.00 & 0.29 \\ \bottomrule
\end{tabular}
\end{table}
\begin{figure*}[ht!]
    \centering
    \begin{subfigure}[t]{0.49\textwidth}
        \centering
        \includegraphics[width=\textwidth]{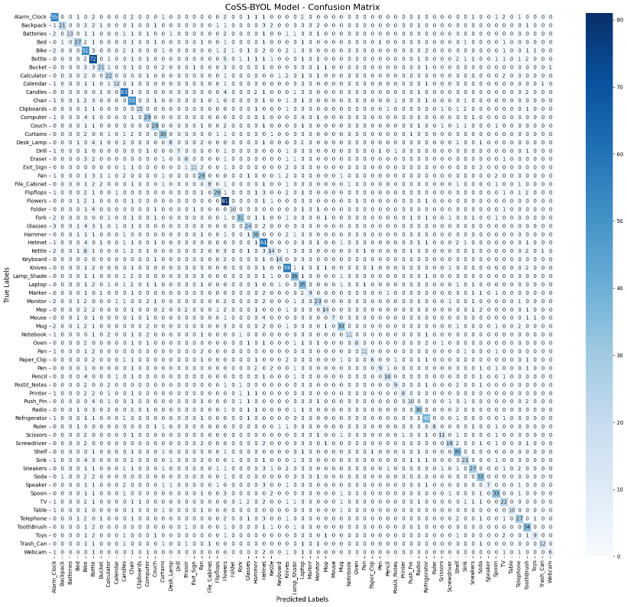}
        \caption{Source domain (Art) after adaptation showing mild diagonal degradation.}
        \label{fig:cm_art}
    \end{subfigure}
    \hfill
    \begin{subfigure}[t]{0.49\textwidth}
        \centering
        \includegraphics[width=\textwidth]{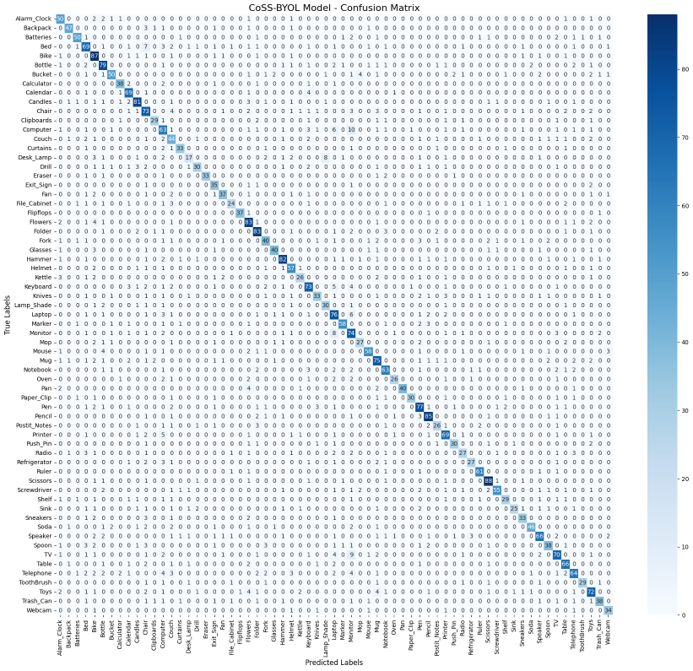}
        \caption{Target domain (Clipart) after adaptation showing strong diagonal concentration.}
        \label{fig:cm_clipart}
    \end{subfigure}
    \caption{Confusion matrices of the SCoDA-adapted BYOL model evaluated on both source and target domains, demonstrating successful knowledge transfer with controlled forgetting.}
    \label{fig:conf_matrices}
\end{figure*}

\subsubsection{Fundamental Challenges with DINO Integration}
We have conducted a further study with the Dino model, as it has shown a slightly different trend in results, compared to the other self-supervised models. Table \ref{tab:dino_performance} shows the performance of DINO-based models on the Office-Home dataset, highlighting the catastrophic forgetting issue. The detailed analysis of DINO's poor performance reveals four critical failure modes that explain why SCoDA struggles with transformer-based SSL models:

\textbf{Destructive Fine-Tuning Dynamics}: SCoDA's aggressive fine-tuning approach fundamentally conflicts with DINO's robust feature learning. The full fine-tuning strategy causes severe feature degradation, evidenced by the dramatic drop in source domain performance from 48.00\% to 4.00\%. This drop in performance suggests that DINO's learned representations are more fragile under aggressive adaptation process (provided by the proposed method) compared to BYOL's more resilient features.

\textbf{Loss Function Incompatibility}: The composite loss design that works well with BYOL and all other CNN based self supervised models, proves counterproductive for DINO. The cosine similarity loss enforces rigid alignment that is overly restrictive for DINO's rich semantic feature space (due to the transformer baseline), while the spatial structure loss introduces instability by conflating semantic differences with domain shift.

\textbf{Representation Collapse}: Unlike BYOL and other CNN based self supervised models, DINO requires specific stabilization mechanisms (centering and sharpening) to maintain meaningful representations. SCoDA lacks these essential components, leading to representation collapse during adaptation.

\textbf{Objective Misalignment}: SCoDA's direct feature alignment approach contradicts DINO's cross-entropy-based learning objective over learned prototypes. This fundamental mismatch prevents SCoDA from leveraging DINO's semantically structured feature space effectively.

To summarize, the proposed SCoDA is expected to perform well for any CNN based self-supervised models. However, for transformer based model, SCoDA needs some modifications as discussed above.

\subsubsection{Qualitative Analysis via Confusion Matrices}
The confusion matrices shown in Figure \ref{fig:conf_matrices} provide confirmation of the adaptation dynamics. The source domain (Art) confusion matrix after adaptation shows a noticeable decline in diagonal dominance, indicating increased misclassification rates and confirming some degree of catastrophic forgetting. However, this forgetting is much less severe than observed with DINO, suggesting that BYOL provides a more stable foundation for adaptation.

Conversely, the target domain (Clipart) confusion matrix demonstrates strong diagonal concentration, indicating improved class separability and successful knowledge transfer. The clear distinction between source and target domain confusion patterns validates that SCoDA successfully transfers discriminative knowledge while maintaining reasonable source domain performance when properly initialized with compatible SSL methods such as BYOL.

\section{Conclusion}
\label{sec:conclusion}
We introduced SCoDA, a framework for unsupervised Source-Free Domain Adaptation. SCoDA departs from the state-of-the-art SFDA paradigm in two fundamental ways. First, we replace the conventional supervised pre-training with a fully self-supervised approach, demonstrating that a label-agnostic initialization scheme provides a more robust and adaptable foundation. Second, we successfully adapt and validate the geometric manifold alignment principle to the SFDA setting. By combining the Space Similarity loss with a stable, dual-speed teacher-student architecture with continual learning setup, SCoDA effectively preserves the holistic structure of the teacher's knowledge while mitigating catastrophic forgetting. Future work could explore the integration of SCoDA's geometric distillation mechanism into other continual or test-time adaptation frameworks. Making the required updatation on SCoDA to make it perform well on transformer based models, may be another possible future research direction.
\bibliographystyle{ACM-Reference-Format}
\bibliography{sample-base}

\end{document}